\def\year{2021}\relax
\documentclass[letterpaper]{article} % DO NOT CHANGE THIS
\usepackage{aaai21}  % DO NOT CHANGE THIS
\usepackage{times}  % DO NOT CHANGE THIS
\usepackage{helvet} % DO NOT CHANGE THIS
\usepackage{courier}  % DO NOT CHANGE THIS
\usepackage[hyphens]{url}  % DO NOT CHANGE THIS
\usepackage{graphicx} % DO NOT CHANGE THIS
\def\UrlFont{\rm}  % DO NOT CHANGE THIS
\usepackage{natbib}  % DO NOT CHANGE THIS AND DO NOT ADD ANY OPTIONS TO IT
\usepackage{caption} % DO NOT CHANGE THIS AND DO NOT ADD ANY OPTIONS TO IT
\usepackage{amsfonts}
\usepackage{amssymb}
\usepackage{amsthm}
\usepackage{amsmath}
\usepackage{tikz}
\usepackage[linesnumbered,noend,ruled]{algorithm2e}
\usepackage{xcolor}

\usepackage{array}
\newcommand{\PreserveBackslash}[1]{\let\temp=\\#1\let\\=\temp}
\newcolumntype{C}[1]{>{\PreserveBackslash\centering}p{#1}}
\newcolumntype{L}[1]{>{\PreserveBackslash\raggedright}p{#1}}

\usepackage{algpseudocode}
\usepackage{comment}

\usetikzlibrary{calc, chains, fit, positioning, shapes, arrows.meta,matrix,%
                decorations.pathreplacing}

\definecolor{myblue}{RGB}{80,80,160}
\definecolor{mygreen}{RGB}{80,160,80}

\usepackage{xcolor}
\usepackage{multirow}

\title{SOLO: Search Online, Learn Offline for Combinatorial Optimization Problems}

\newcommand{\note}[2]{}
\newcommand\dd[1]{ \note{magenta}{[DD: #1]}}

\newcommand\fr[1]{ \note{red}{[FR: #1]}}
\newcommand\jo[1]{ \note{blue}{[JO: #1]}}

\newcommand\chr[1]{ \note{orange}{[CR: #1]}}

\author{
    Joel Oren$^1$, Chana Ross$^1$, Maksym Lefarov$^1$, Felix Richter$^1$, Ayal Taitler$^2$, Zohar Feldman$^1$, \\
    Dotan Di~Castro$^1$, Christian Daniel$^1$\\
}
\affiliations{
    \textsuperscript{\rm 1}Bosch Center for AI \\
    \{Joel.Oren,chana.ross,zohar.feldman,dotan.dicastro\}@il.bosch.com,
    \{maksym.lefarov,felixmilo.richter,christian.daniel\}@de.bosch.com \\ 
    \vspace{1mm}
    \textsuperscript{\rm 2}Technion -- Israel Institute of Technology\\
    ataitler@technion.ac.il

}

\begin{document}

\maketitle

\begin{abstract}

We study combinatorial problems with real world applications such as machine scheduling, routing, and assignment. We propose a method that combines Reinforcement Learning (RL) and planning. This method can equally be applied to both the offline, as well as online, variants of the combinatorial problem, in which the problem components (e.g., jobs in scheduling problems) are not known in advance, but rather arrive during the decision-making process. Our solution is quite generic, scalable, and leverages distributional knowledge of the problem parameters.
We frame the solution process as an MDP, and take a Deep Q-Learning approach wherein states are represented as 
graphs, thereby allowing our trained policies to deal with arbitrary changes in a principled manner.
Though learned policies work well in expectation, small deviations can have substantial negative effects in combinatorial settings. We mitigate these drawbacks by employing our graph-convolutional policies as non-optimal heuristics in a compatible search algorithm, Monte Carlo Tree Search, to significantly improve overall performance. We demonstrate our method on two problems: Machine Scheduling and Capacitated Vehicle Routing. We show that our method outperforms custom-tailored mathematical solvers, state of the art learning-based algorithms, and common heuristics, both in computation time and performance.

\end{abstract}

\section{Introduction}\label{sec:introduction}
Combinatorial optimization (CO) is a central area of study in computer science with a vast body of work that has been done over the past several decades.
Despite growing compute resources and efficient solvers, many practical problems are computationally intractable, and therefore, problem-specific heuristics and approximation methods have been developed \cite{williamson2011design}.
However, these  methods suffer from two limitations. First, they are usually highly specialized: they require the algorithm designer to gain insights about the problem at hand. Second, they typically aim for worst-case scenarios; this often renders them less useful in practical applications, where average-case performance is key.

Researchers have recently studied the applicability and advantages of Reinforcement Learning (RL) methods for CO problems. In particular, and unlike analytical methods, learned policies can be evaluated in near real-time once trained, thus enabling fast response times without sacrificing the solution quality. While early work concentrated on simplified state-action spaces (e.g., when to run a given heuristic in a search procedure \citealt{Khalil2017, kruber2017learning}), recent advances in RL have enabled the training of policies on more expressive representations of the state-action space (see e.g., \citealp{Waschneck2018a}).

A common RL approach to CO is to incrementally construct solutions by treating the partial solutions as the \textit{states} and their extensions as \textit{actions}. We propose to model the state-action space as a \textit{graph} and learn a graph neural network (GNN) policy that operates on it. GNNs have recently become a popular neural method for learning representations of rich combinatorial structures using graphs. They offer a number of desireable properties such as invariance to node permutation and independence of graph and node neighborhood sizes. Specifically in our case, they allow us to handle problems of different sizes in states and actions spaces, using the same compact network. We represent each \textit{separate} observation as a graph, which lets us deal with dynamic changes in the problem. We then extend existing Deep Q-learning approaches (DQN; \citealp{mnih2013playing, hessel2018rainbow}) to learn effective policies that are agnostic to the number of components in the problem instance (e.g., locations in routing problems, jobs in scheduling problems), and compatible with online stochastic arrivals. We train our GNN models to predict the Q-values of states, by making use of their applicability to graphs of varying sizes. We note at this point that exploring GNN architectures was not the main focus of this work, but utilizing a common architecture and in our planning-learning scheme.

% A limiting factor to the above approach is the fact that c
Combinatorial problems are notorious for being sensitive to slight perturbations in their solutions (see e.g., \citealp{hall04schedulingsensitivity} and Chapter 26 in \citealp{gonzales07handbookapprox}). Drawing inspiration from recent self-play methods for board games (\citealp{silver2017mastering}), we complement our policy with a search procedure, using our trained Q-network as a guide in a modified Monte-Carlo Tree Search (MCTS).
We demonstrate the efficacy of our hybrid method, on two NP-hard problems \cite{KRAMER2019,toth2014vehicle} and their \textit{online} variants: the Parallel Machine Job Scheduling problem (PMSP), and the Capacitated Vehicle Routing problem (CVRP).

%we mitigate degradation in performance by complementing our policy with a search procedure. We do so by using our trained Q-network as a guide in a modified Monte-Carlo Tree Search procedure.

\vspace{0.5mm}
\textbf{Contributions:} 
We propose a method that incorporates end-to-end GNN models. This allows our trained, fixed-size policies to operate on problems with varying state and action sizes. In particular, our approach is capable of solving \textit{offline} and \textit{online} problems of magnitudes that are not known a priori. To mitigate degradation in performance of learned solutions, we devise a scheme that bridges the gap between searching online and learning offline, named \textit{SOLO}. This approach benefits from the generalization capacity, and real-time performance of trained policies, as well as the robustness obtained by online search. Our results show that SOLO finds higher quality solutions than existing learning-based approaches and heuristics, while being competitive with fine-tuned analytical solvers.

% compare SOLO to recent learning-based approaches as well as to common heuristics. Our results show that it finds higher quality solutions than existing learning-based approaches and heuristics, while being competitive with fine-tuned analytical solvers.
% We demonstrate the efficacy of our method, \textit{SOLO}, on two NP-hard problems \cite{KRAMER2019,toth2014vehicle} and their \textit{online} variants: the \textit{Parallel Machine Job Scheduling} problem (PMSP), and the \textit{Capacitated Vehicle Routing} problem (CVRP).

\section{Related Work}\label{sec:related_work}

The application of Machine Learning (ML) approaches to NP-hard problems \cite{garey90npcompleteness}
%, williamson2011design} 
has gained much interest recently. While they are generally hard to solve at scale, these problems are amenable to approximate solutions via ML algorithms \cite{Bengio2018}. In general, there are three popular approaches for solving NP-Hard combinatorial problems.  The most basic one includes non ML algorithms such as branch \& bound search.
Another, applies ML techniques in an end to end manner. 
Finally, there are works that combine the two previous methods together in an augmented way.

\vspace{0.5mm}
{\bf Classical Search Methods:} Researchers in Operations Research and Computer Science have developed an extensive set of tools and algorithms for CO problems over the years \cite{korte2011combinatorial}.
%borodin2005online,williamson2011design}. 
Their drawbacks are usually the need for domain-specific tailoring or extensive computational requirements \cite{wolsey1999integer}. 
\citealp{peters2019mixed} compared local search via genetic algorithms and Integer Programming in a staff assignment problem, and showed that local search can produce sub-optimal solutions relatively fast. 
Many other local search methods using neighborhood relations have been established, and applied to problems such as paint shop scheduling \cite{winter2019solution}, scheduling with time widows \cite{he2019tabu} and more. 
%\ml{guys, I admit I've never written a paper, but I'm not sure that "celebrated" algorithm, method etc. is an appropriate phrasing for the scientific paper} 
% MCTS algorithm which applies Upper Confidence Bounds to Trees \cite{kocsis2006bandit} to prune them during search and %achieve optimal policies faster
A popular approach for global search is Monte-Carlo Tree Search (MCTS). The Upper Confidence-Bounds applied to Trees (UCT) \cite{kocsis2006bandit} is a particular instance of MCTS, which provides formal guarantees on optimality and achieves attractive results in practice. Distributed approaches have also been investigated by the community, \citealp{Nicolo191911_search} solved a multi-agent scheduling problem using a distributed structure. They have allocated different agents that execute simultaneous local searches on previously found solutions.

\vspace{0.5mm}
\textbf{End To End Machine Learning:} Common approaches apply RL algorithms in order to generate solutions without the use of problem specific heuristics. \citealp{KhalilEtAlNIPS2017} implemented a combined DQN and GNN architecture in order to learn a greedy search for graph optimization problems.
\citealp{bello2016neural} used an RNN and REINFORCE \cite{williams1992simple} to train an agent to solve the Travelling Salesperson problem (TSP). 
\citealp{NIPS2018_nazari} employed a similar method with the addition of an attention mechanism. \citealp{Kool2019ICLR} also approached routing problems using a GNN architecture combined with an attention encoder-decoder mechanism and the REINFORCE algorithm. These three works employ a \textit{masking} step on their outputs to disallow unavailable actions (e.g., locations already visited). This approach requires the set of actions to be set in advance.
% \citealp{Mao2016,MaoEtAl2019}
\citealp{MaoEtAl2019}
also used REINFORCE with graph convolutional embedding for job scheduling on a cluster, but in an online setting where the jobs had a DAG structure that could be exploited. No search was incorporated in here due to the real-time dynamic nature of the processing cluster scheduling. Recently \citealt{joe2020deep} applied DQN in the Dynamic Vehicle Routing Problem to learn approximated value function and a routing heuristic.

\vspace{0.5mm}
\textbf{Combined Search and Machine Learning:}
Combination of search algorithms and ML techniques can benefit one another by either employing search to accelerate the learning, or learning better models for the search to use or both.
% When combined, search algorithms and ML techniques can benefit one another by either (a) using the search algorithms to accelerate the learning process, or (b) use learned models for achieving faster results in the search tree, or (c) a combination of the two.  
\citealp{Waschneck2018a} applied RL methods for optimizing scheduling problems with a multi-agent cooperative approach. However, their experiments did not demonstrate clear improvements over heuristic algorithms. \citealp{chen2019learning} took a different approach, they used a DQN for choosing regions in the solution to improve a local search heuristic. We use their method as a baseline in our experiments section.
\citealp{Zhuwen2018} used GNN and supervised learning to label nodes in a graph, to determine whether they belong to a Maximal Independent Set. This prediction was used within a tree search algorithm to find the best feasible solution the network predicted.
Recently, MCTS combined with RL methods has gained much traction due to superhuman play level in board games such as Go, Chess, and Shogi \cite{silver2017mastering}. \citealp{instadeep_ranked_rewards18} have integrated MCTS into a RL loop and solved the Bin Packing Problem (BPP) in two and three dimensions using ranking rewards.

% \textbf{Our Approach} We use the combined approach of integrating DQN with MCTS, allowing the search algorithm to benefit from the learned model. Our solution for the CO problems first learns the Q-values offline using a GNN approximation and a variant of the DQN algorithm (we name this model Q-net). Once the model has trained, we use MCTS online and find the optimal solution guiding the search with our Q-net.
% Many previous RL works required for fix length state and action spaces representation throughout the episodes. Others used graph representations, but maintained the action space fixed with network masking. Instead we allow for changing spaces length by constructing a new graph representation at each step.
% allowing for compatibility with online computation settings \cite{borodin2005online}. 

% that their inputs (that is, the graph representations of the states) needed to remain fixed throughout the episodes. Some incorporated a final fully-connected layer in their architectures. Others used graph representations throughout, but maintain the action space fixed, while applying masking on the network outputs. Instead, we construct a new graph at each step, allowing for compatibility with online computation settings \cite{borodin2005online}. 

\section{Method and Definitions}
\subsection{Modeling Combinatorial Optimization Problems}
A combinatorial optimization (CO) problem is given by a triplet $\big< \mathcal{I}, S, f \big>$, where $\mathcal{I}$ is the set of problem instances, $S$ maps an instance $I \in \mathcal{I}$ to its set of feasible solutions, and $f$ is the objective function mapping solutions in $S(I)$ to real values. We model a sequential solution process of an instance $I$, in which at each time $t$ a partial solution is extended,  using a finite horizon Markov Decision Process (MDP; \citealp{puterman1994markov}) of $T$ steps. At each time $t$, the state $s_t$ corresponds to a partial solution, an action $a_t$ corresponds to a feasible extension of $s_{t}$ (similarly to a greedy algorithm), a reward  $r_{t+1} = r(s_t,a_t) = f(s_{t+1}) - f(s_t)$,\footnote{Note that a different reward function that equals the resulting solution value at the end of the episode, and zero anywhere else would tend to make the learning process harder, due to its sparsity.} and a transition probability  $p(s'|s_t,a)$. The action distribution is set by a policy $\pi(a|s)$. This leads to a distribution over \textit{trajectories} $\rho = (\left< s_t, a_t, r_{t+1} \right>)_{t=0,\ldots,T-1}$, $p(\rho)= p(s_0)\prod_{t=0}^{T-1}\pi(a_t|s_t)p(s_{t+1}|s_t,a_t)$.
The $Q$-function is defined as
\begin{align*}
    Q(s_t, a_t) \triangleq \mathbb{E}_{\rho}\left[\left. \sum_{i=0}^{N-t}  r(s_i,a_i)\right|s_0=s_t, a_0=a_t\right],
\end{align*}
where the agent's objective is to find an  optimal policy  $\pi^*(a|s)= \arg \max_a Q(s,a)$ \cite{sutton2018reinforcement}.

We encode a problem instance $I$, or rather, a sub-problem thereof, induced by a partial solution $s_t$, as a graph $G_{t}=(V_t, E_t, f^v_t, f^e_t)$ where $f^v_t$ (resp. $f^e_t$) maps nodes (edges) to feature vectors. This encoding is often quite expressive and allows for a systematic treatment of many CO problems.

\subsection{Motivating Combinatorial Problems}
\label{sec:motivating_problems}
Our approach is quite generic and can be applied to many optimization problems. We illustrate its efficacy on two well-known problems: the Parallel Machine Scheduling Problem \cite{KRAMER2019}, or PMSP for short, and a simplified version of the Capacitated Vehicle Routing problem (CVRP), introduced by \citealp{dantzig59}. Both PMSP and the CVRP are widely-known problems, however, we briefly outline them here.

CVRP is widely known, but briefly: there is a  set $N = \{ 1,\ldots,n \}$ of customers with demands $\{d_i \}_{i \in N}$ for a single commodity, and locations $\{p_i \}_{i \in N}$. The commodity is located at a depot $o$, and is to be transported from it to the customers so as to meet their demand by a vehicle of capacity $C^*$ (for feasibility, we assume that $d_i \leq C^*$ for all $i \in N$) and velocity $\nu$. A \textit{tour} consists of visited locations,  $\tau = (p_1 = o, \ldots, p_{n_{\tau}} = o)$, in which the vehicle's load does not exceed its capacity and satisfies the demands of the customers along it. The objective is to minimize the total distance traveled by the vehicle.

PMSP is a scheduling problem with $m$ (unrelated) machines, and $n$ jobs with weights $\big( w_i \big)_{i=1,\ldots,n}$ and processing times $\big( p_i \big)_{i=1,\ldots,n}$. Also, each job $u$ belongs to a job class $\kappa_u \in \{1,\ldots,c\}$, so that if job $u$ is scheduled to process immediately after job $u'$ on the \textit{same} machine, then there is an additional incurred \textit{setup time} $P[\kappa_{u'}, \kappa_u]$, where $P$ is a matrix with non-negative entries, and zeros in the diagonal. The objective is to minimize the total \textit{weighted} completion time, which consists of the waiting, setup, and processing times of all arrived jobs.  \fr{This is not quite correct: we used the total weighted completion time; so it's weighted, and it's completion time instead of processing time. Note that completion time is different from processing time, the completion time of a job is waiting time + processing time + setup time} \dd{Do we want to emphaisze that if a class doesn't change then $P[\kappa_{u'}, \kappa_u] \ge 0$ ?} \jo{I added a quick sentence about the setup time matrix, and changed processing time to completion time.}

\vspace{0.5mm}
\textbf{Online Variants:}
%\paragraph{Online Variants}
Both problems have online variations, in which the problem constituents are \textit{not} known apriori, but rather, arrive according to certain distributions, at the time in which a solution is constructed. In the case of online CVRP, customers arrive at different times, as the vehicle proceeds along its current route. In PMSP, jobs appear as machines are processing the previously arrived jobs.

\section{Solution Approach}\label{sec:solution_approach}
\begin{figure*}
\includegraphics[width=\linewidth]{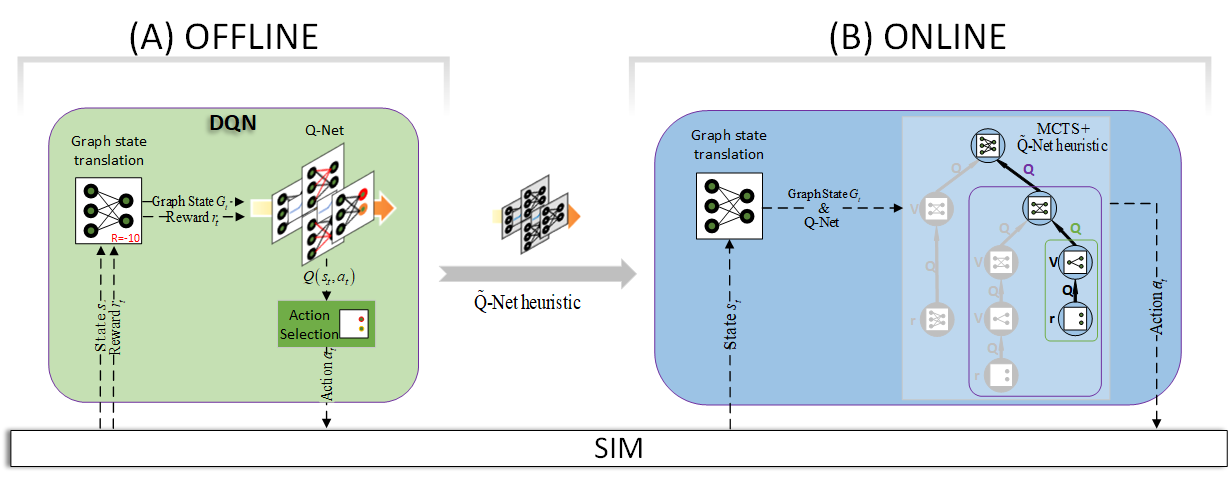}
\caption{A schematic overview of SOLO. On the left, a depiction of our DQN training process, which produces the $\tilde Q$-Net heuristic. On the right is our planning procedure that, for each step, runs our modified MCTS with $\tilde Q$-Net as a heuristic.\jo{Added a real caption here.}}
\label{fig:method}
\end{figure*}
Our solution is composed of two phases as depicted in Figure~\ref{fig:method}. First, we train an agent offline, using a simulated environment, in an off-policy manner, where representations of the states are learned using a graph neural network. Second, at solution time, we use the policy trained offline as a heuristic in a Monte-Carlo tree search (MCTS). This approach allows us to train offline a good policy, while the online tree search ensures a more robust solution.

\vspace{0.5mm}
% \paragraph{Learn Offline}
\textbf{Learn Offline:}
We base our offline part of the method on the off-policy DQN algorithm where we learn a neural approximation $\tilde{Q}(s,a;\theta) \approx Q(s, a)$, parametrized by $\theta$ \cite{mnih2013playing}. 
Modelling $\tilde{Q}$ using feed-forward or convolutional neural network architectures is challenging in the CO setting as the sizes of the state and action spaces can vary throughout an episode. For this reason, we encode a state-action tuple as a graph $G_t = (V_t, E_t, f^v_t, f^e_t)$ with sets of nodes $V_t$, edges $E_t$, node features $f^v_t$ and edge features $f^e_t$.
See the next section
% section~\ref{sec:implementation_details} 
for problem-specific descriptions of the graph encodings.

As shown in Figure~\ref{fig:method}, we train our $\tilde{Q}$ model using a simulation of the problem. At decision time $t$, the current state $s_t$ is translated to a graph, and the next action is chosen using the model's prediction of the $Q$-values for each state-action tuple $(s_t, a_t)$. We take an epsilon-greedy approach in the training phase for choosing the next action.
We model $\tilde{Q}$ as a Graph Neural Network (GNN) with the following three components: (1) embedding model: a vanilla feed forward model with leaky \textit{ReLU} activation function. This component converts the features of each node to a higher dimension. (2) encoder model. (3) decoder model. The encoder and decoder models both use a variant of the message-passing GNN architecture \cite{battaglia2018relational} with the following update rules:
\begin{equation*}
    \begin{split}
        e_i^{(t+1)} & = \phi_e(e_i^{(t)}, \rho^{v\to e}(V), w), \forall e_i \in E,\\
        v_i^{(t+1)} & = \phi_v(v_i^{(t)}, \rho^{(e, v)\to v}(V, E), w), \forall v_i \in V,\\
        w^{(t+1)} & = \phi_w(w^{(t)}, \rho^{(e, v)\to w}(V, E)),
    \end{split}
\end{equation*}
Where $\rho^{v\to e}$, $\rho^{(e, v)\to v}$, $\rho^{(e, v)\to w}$ are the aggregation functions that collect information from the \textit{related} parts of $G_t$, and $\phi_e$, $\phi_v$, $\phi_w$ are the neural networks that update the feature vectors $f^v_t$, $f^e_t$ and $w$, respectively ($w$ being a global feature of the graph).

Each neural network consists of a number of fully connected layers (three layers in the encoder model and one in the decoder model), followed by a leaky \textit{ReLU} activation function and finally a normalization layer. Our DQN training process includes several common techniques used in deep Q-learning, such as double DQN and priority replay (see e.g., \citealp{hessel2018rainbow}). More information about our model can be found in the supplementary material.

We compute $\tilde{Q}(s, a)$ as the the output of $\phi_e$ at the final message-passing step of the decoder model since the edges represent the actions in our problems.
This architecture allows us to address the changing size of the state (i.e., different $|V|$ and $|E|$ per step) in a principled manner. A nice property of GNN's is that the results are insensitive to ordering in the input, making them particularly effective for un-ordered data such as sets and graphs.

\vspace{0.5mm}
\textbf{Search Online:}
% \paragraph{Search Online} 
At run time, we use online search to optimize the decisions in the states observed in real time. Namely, we take the time until the next decision is due (e.g., the next job assignment or the next customer pick-up) to evaluate all the applicable actions in the current state, and upon  reaching the timeout, output the estimated best action for this state.
We employ a modified version of the popular MCTS algorithm UCT (Upper Confidence-Bound applied to Trees, \citealp{kocsis2006bandit}) in conjunction with the $\tilde{Q}$-value that was trained offline. The algorithm works by sampling rollouts iteratively from the current root state, $s_t$, until we reach a terminal state. After each iteration of the online search, the $Q$-value of any previously encountered state-action pair $(s,a)$ is estimated as $Q^{\mathcal{T}}\left(s,a\right)=\sum_{i=1}^{n(s,a)}  \frac{\overline{r}_i}{n(s,a)}$, with $n(s,a)$ denoting the number of times action $a$ was sampled in state $s$ over all the sampled rollouts. Each reward-to-go $\overline{r}_i$ is the accumulated reward from state $s$ onward in a corresponding rollout that passed through $(s, a)$. Similarly to UCT, we select actions during each rollout according to the Upper Confidence-Bound (UCB) policy. Namely, we select an action that was not yet sampled in the current state, if such action exists, and otherwise we select the action that maximizes the value $Q^{\mathcal{T}}\left(s,a\right) + \beta\sqrt{\frac{\log\left(n(s)\right)}{n(s,a)}}$, with $n(s)=\sum_{a}n(s,a)$, and $\beta$ is the exploration factor that governs the balance between exploration and exploitation. Algorithm~\ref{alg:mcts} depicts our online search. 
While UCT is guaranteed to asymptotically converge at the optimal decision by sampling each action infinitely often, in order to improve its practical efficiency, we combine it with the following features:

\textit{MCTS heuristic:} 
We use the $\tilde{Q}$-value as an out-of-tree heuristic. We employ an iterative deepening scheme according to which, in each iteration, the search tree is expanded with at most one state, which is the first state in the rollout that is not already in the tree. During a rollout, whenever we encounter states that are not yet in the tree, we choose the action that maximizes the $\tilde{Q}$-value rather than using UCB (corresponding to a random action in this case). This is a common technique that often improves performance significantly, depending on the quality of the used heuristic.

\textit{Action pruning:} Second, we use the $\tilde{Q}$-values to \textit{prune the action space}. Since the number of plausible actions in CO problems tends to be intractably high, we would like to focus the samples on actions that have higher potential to be a (near-)optimal action. We thus consider for each state $s$ only the $k$ actions with the highest $\tilde{Q}$-value, where $k$ is a hyper-parameter of our algorithm. Indeed, as the estimated $\tilde{Q}$-value becomes more accurate, the probability that one of these $k$ actions is near-optimal increases.

\textit{Rollout preemption} Additionally, to allow for more sampled rollouts, we shorten the lookahead period by suppressing future arrivals (e.g., arrivals of jobs or customers) that are scheduled for a time after $t+\Delta_T$, where $\Delta_T$ is a hyperparameter). As this technique is mostly implemented in the simulation, it is not depicted in Algorithm~\ref{alg:mcts}. By applying the above features, we theoretically compromise optimality, but our experiments demonstrate that they hold some practical merits.

\section{Implementation Details}
\label{sec:implementation_details}
In the following section we show how to represent states, actions, and rewards in our solution framework.   

\vspace{0.5mm}
\textbf{Graph States:}
% \paragraph{Graph States}
In \textit{CVRP} a state $s_t$ is given by the set of $n_t$ pending customers, their coordinates $\mathbf{p_i} \in \mathbb{R}^2$ and demands $d_i \geq 0$, for $i=1,\ldots,n_t$. The current state also specifies the current vehicle location $\mathbf{p^*}$, along with its remaining capacity $c^*$, and the location of the depot $\mathbf{p_o}$. To encode $s_t$, we use a simple star graph topology with $n_t+2$ nodes that consist of a vehicle node $u^*$ as the central node that is connected to $n_t$ customer nodes $\{u_{i}\}_{i=1,\ldots,n_t}$ and a depot node $u_o$. We employ a unified node feature vector that holds entries for the features of all three node types (with zeros where inapplicable): node $u$'s location $p_u$, capacity $c^*$ (vehicle), and demand (customers). A node feature vector also includes a length-$3$ one-hot encoding denoting the node type. Finally, a node $u_i$'s feature vector contains a binary value $\mathbb{I}[d_i \leq c^*]$, indicating (in the case of a customer) whether the vehicle can travel to it. As for edges connecting the customer to the depot and vehicle, their features consist of solely their respective distances.

In \textit{PMSP}, a state $s_t$ consists of the currently pending $n_t$ jobs, the set of $m$ machines along with the remaining processing times $(r^{(i)}_t)_{i=1,\ldots,m}$ and their last or currently assigned job classes (for computing setup times), $(\kappa^{last}_{t,i})_{i=1,\ldots,m}$. We represent this as a complete bipartite graph $(V_t=(J_t,M), E_t, f_t^{v}, f_t^{e})$, where $J_t$ and $M$ are the sets of job and machine nodes, respectively, $E_t=J_t \times M$ is the complete set of edges, connecting every job to every machine. and lastly, $f_t^v$ and $f_t^e$ map nodes and edges to feature vectors. For a job node $u_j \in V_t, j=1,\ldots,n_t$, its feature vector $f^{v}_t(u_j)$ consists of its processing time $p_{j}$, weight $w_j$, and arrival time $a_j$ (positive in the online case), a length-$c$ one-hot encoding indicating the job class $\kappa_{j}$, and a node-job indicator $\mathbb{I}^{job}_{u_j} = 1$. For a machine node $v_{i} \in M, i=1,\ldots, m$, in addition to a machine indicator bit $\mathbb{I}^{job}_{v_i} = 0$, we also encode the remaining processing time of the machine $r^{(i)}_t$, and a one-hot encoding of length $c+1$ specifying the class of last run job, $\kappa^{last}_{t,i}$ (including the special ``empty class'' for machines that were not yet assigned jobs). For an edge $e_{ji}$, connecting job node $u_j$ to machine node $v_i$, the only feature is the incurred setup time $P[\kappa_{j}, \kappa^{last}_{t,i}]$.\footnote{If machine $i$ was not previously assigned a job, than the setup time would be zero.} As before, we keep the node feature vectors at equal lengths, by having every node vector contain entries for both job and machine features; for job nodes the machine feature entries are zeroed out, and vice versa.
Figures \ref{fig:pmsp_bipartite} and \ref{fig:pmsp_features} depict the resulting complete bipartite graph and feature vectors, respectively.
\begin{figure*}[ht]
%\centering
\begin{minipage}[b]{.4\textwidth}
\centering
\begin{tikzpicture}[
node distance = 4mm and 21mm,
  start chain = going below,
     V/.style = {circle, draw, 
                 fill=#1, 
                 inner sep=0pt, minimum size=3mm,
                 node contents={}},
every fit/.style = {ellipse, draw=#1, inner ysep=-1mm, 
                 inner xsep=5mm},
                ]
% vertices 
\node (n11) [V=myblue,on chain,
              label={[text=myblue]left:$u_{1}$}];
\node (n21) [V=mygreen, right=21mm of n11,
              label={[text=mygreen]right:$v_{1}$}];

\node (n12) [V=myblue,on chain,
              label={[text=myblue]left:$u_{2}$}];
\node (n22) [V=mygreen, right=21mm of n12,
              label={[text=mygreen]right:$v_{2}$}];

\node (n13) [V=myblue,on chain,
              label={[text=myblue]left:$u_{3}$}];
\node (n23) [V=mygreen, right=21mm of n13,
              label={[text=mygreen]right:$v_{m}$}];
\node (n14) [V=myblue,on chain,
              label={[text=myblue]left:$u_{n_t}$}];
% set U
\node [myblue,fit=(n11) (n14),label=above:$J_t$] {};
% set V
\node [mygreen,fit=(n21) (n23),label=above:$M$] {};
% links
\draw[-, shorten >=1mm, shorten <=1mm]
    (n14) edge (n23)   (n12) edge (n23)
    (n13) edge (n23)   (n13) edge (n22)
    (n13)
    (n12) edge (n21)   (n12) edge (n21)
    (n11) edge (n22)   (n12) edge (n22)
    (n11) edge (n23)   (n11) edge (n21);

\path (n13) -- (n14) node [red, midway, sloped] {$\dots$};
\path (n22) -- (n23) node [red, midway, sloped] {$\dots$};
\end{tikzpicture}
\caption{PMS: The GNN representation of a state $S_t$ with $n_t$ waiting jobs and $m$ machines.}
\label{fig:pmsp_bipartite}
\end{minipage}
\qquad\qquad\qquad
\begin{minipage}[b]{.45\linewidth}
\begin{tikzpicture}[
    MyStyle/.style={draw, minimum height=2em, minimum width=2.9em, outer sep=0pt},
  ]
% Adapted from https://tex.stackexchange.com/a/246976
\matrix (A) [matrix of math nodes, nodes={MyStyle, anchor=center}, column sep=-\pgflinewidth]
{p_j & w_j & a_j & \mathbf{h}_{\kappa_{u_j}}^{c} & \mathbb{I}^{node}_{u_j} & & 0 & 0 \\};

\begin{scope}[yshift=-1cm]
\matrix (B) [matrix of math nodes, nodes={MyStyle, anchor=center}, column sep=-\pgflinewidth]
{0 & 0 & 0 & 0 &  \mathbb{I}^{node}_{v_i} & r_i & \mathbf{h}_{\kappa_{v_i}}^{c+1}\\};
\draw[decorate,decoration={brace, amplitude=10pt, raise=5pt}]
  (A-1-1.north west) to node[black,above,above= 15pt] {job features} (A-1-4.north east);%
\draw[decorate,decoration={brace, amplitude=10pt, raise=5pt}]
  (A-1-7.north west) to node[black,above,above= 15pt] {machine features} (A-1-8.north east);%
\end{scope}
\end{tikzpicture}
\caption{Feature vectors for job node $u_j$ and machine node $v_i$ (PMSP).}
\label{fig:pmsp_features}
\end{minipage}
\end{figure*}

% = \left( (\mathbf{p_i})_{i=1,\ldots,n_t}, (\mathbf{d_i})_{i=1,\ldots,n_t}, \textbf{p^*}, c^*, \textbf{p_o} \right)

\vspace{0.5mm}
\textbf{Actions:}
% \paragraph{Actions}
In both settings, the set of actions are specified by the set of edges in the graph representation of the current state $s_t$: in CVRP, selecting an edge between $u^*$ and $u_i$ (resp. $u_o$), corresponds to extending the route traveled thus far by instructing the vehicle to drive to a customer (resp. depot) in location $\mathbf{p}_i$ (resp. $\mathbf{p_o}$). Note that an edge to a customer node $u_i$ is \textit{feasible} provided that $d_i \leq c^*$. Similarly in PMSP, the set of actions is specified by the edges connecting jobs to machines. Selecting an edge $(u_j, v_i)$ corresponds to assigning job $j$ to machine $i$ at the time of state $s_t$. Importantly, in both settings we also allow for the special \textit{noop} action that simply has no effect apart from ``skipping'' to the next decision point (see below). Additionally, a given state might induce a graph representation with certain disallowed edge actions. Though one can simply remove these edges from the graph, we decided to leave them in, and apply action masking. We found this approach to be more effective since it enables better flow of information between the nodes, and gave overall better results.

{\SetAlgoNoLine
\begin{algorithm}[!t]
\caption{MCTS with pruned action search}\label{alg:mcts}
    \DontPrintSemicolon
    \SetKwFunction{Fupdate}{update\_tree}
    \SetKwFunction{Ftop}{top\_actions}
    \KwIn{State $S^{t}$, Environment $env$, $Q$-value estimator $\tilde Q$, random seed $seed$, rollout budget $r\_max$, time budget $t\_max$, prune limit $k$, state visits counter $n_s$, state-action visits counter $n_{sa}$, Search Tree estimator $Q^{\mathcal{T}}$}
    env.set\_seed(seed) \\
    
    \For{$r \leftarrow 1,\ldots,r\_max$}
    {
        \If{elapsed\_time() $> t\_{max}$}{
            break
        }  
        $s \leftarrow$ env.set\_state$(S^t)$ \\
        $done, \rho \leftarrow false, [] $\\
        $t_{out}\leftarrow\infty$\\
        $t\leftarrow 0$\\
        \While{not done}
        {
            $t\leftarrow t+1$\\
            $\tilde{q}\leftarrow\tilde{Q}(s,\cdot)$\\
            %$top\_Q \leftarrow$ get\_top\_actions$(s, \tilde{Q}, k)$ \\
            \If{$n_s\left(s\right) == 0$}{
                \tcp*{action with max. Q-value}
                $a \leftarrow \arg_a\max \left(\tilde{q}[a]\right)$\\
                $t_{out}\leftarrow \min(t, t_{out})$
            }
            \Else{
                $top\_actions\leftarrow \mathrm{top\_k\_args}_a(\tilde{q}[a], k)$\\
                $q_{UCB} \leftarrow \{\}$ \\
                \For{$a \in top\_actions$}{
                    $q_{UCB}\left[a\right] \leftarrow Q^{\mathcal{T}}(s, a) + \beta \sqrt{\frac{\log{n_s(s)}}{n_{sa}(s,a)}}$
                }
                $a \leftarrow \arg_a\max \left(q_{UCB}[a]\right)$ \\
            }
            $r, s', done  \leftarrow$ env.step$(a)$ \\
            $\rho\left[t\right]\leftarrow\left< s, a, r \right>$ \\
            $s \leftarrow s'$ \\
            
        }
        update\_tree$(\rho, n_s, n_{sa}, Q^{\mathcal{T}}, t_{out})$
    }
  \Return $\arg_a\max \left(Q^{\mathcal{T}}\left(S^t, a\right)\right)$  \\
%   \SetKwProg{Fn}{Def}{:}{}
%   \Fn{\Ftop{s, \tilde{Q}, k}}{ \\
%             $Q \leftarrow \tilde{Q}(s)$  \\
%             \KwRet{top-$k$ tuples in $Q$ by their value estimates.}}
  \SetKwProg{Fn}{Def}{:}{}
  \Fn{\Fupdate{$\rho, n_s, n_{sa}, Q^{\mathcal{T}}, t_{out}$}}{ 
    $\overline{r} \leftarrow 0$ \\
    \For{$i \leftarrow \left|\rho\right| \text{ to } 1$}{
        $s, a, r \leftarrow \rho[i]$ \\
        $\overline{r} \leftarrow r + \gamma \overline{r}$ \\
        \If{$i>t_{out}$}{continue}
        $n_s\left(s\right)\leftarrow n_s\left(s\right)+1$ \\
        $n_{sa}\left(s,a\right)\leftarrow n_{sa}\left(s,a\right)+1$ \\
        $Q^{\mathcal{T}}\left(s, a\right) \leftarrow \frac{n_{sa}\left(s,a\right)-1}{n_{sa}\left(s,a\right)}Q^{\mathcal{T}}\left(s, a\right) + \frac{1}{n_{sa}\left(s,a\right)} \overline{r}$
    }}
\end{algorithm}
}

\vspace{0.5mm}
\textbf{Decision Events:}
In CVRP a decision point at time $t$ occurs when the vehicle has reached its previous destination, or in the online setting where a \textit{noop} action was previously taken and a new customer arrives. In PMSP, a decision point at time $t$ occurs when a machine has become free and there is at least one pending job, or conversely: a new job has arrived since the previous decision point (in the online setting), and there is a free machine.

\vspace{0.5mm}
\textbf{Reward Function:}
Both CVRP and PMSP are minimization problems, and hence each time $t$ would incur a cost, or a negative reward, taken to be the difference in the objective value for the partial solution resulting from extending the current partial solution by taking $a_t$. In CVRP, this translates to simply the distance traveled between step $t$ and $t+1$. In PMSP, it is the product of the total weight of the jobs processed between steps $t$ and $t+1$ and their time difference.

\section{Empirical Evaluation}\label{sec:empirical_evalutaion}
We evaluate SOLO on the online and offline variants of the two problems mentioned. With minor changes in the graph representation and small hyper-parameter tuning, we manage to achieve competitive results in the online setting without degrading the offline results.

\subsection{Experimental Setup}
As mentioned in previous sections all problems are represented as a bipartite graph (in CVRP we have a simplified star graph) where the actions are edges and the agents or assignments are nodes.
We compare all problems to known baselines including naive solutions, optimal solutions (CPLEX or OR-Tools), problem tailored heuristics and a learned RL solution (Neural Rewriter, \citeauthor{chen2019learning}). More information about the baselines can be found in the supplementary material. 
For fair comparison, we use the same hardware and training time for all networks trained (our Q-net and Neural Rewriter).\footnote{$18$-core Intel Xeon Gold $6150$ @ $2.70$ GHz, Nvidia Tesla V$100$ ($16$ GB)}\footnote{all policies were trained for less than 24 hours} In addition, MCTS and optimization algorithms are given $10$ seconds to optimize each decision. 

The known algorithms for online problems are typically designed to minimize the competitive ratio, i.e., the worst-case ratio of the an algorithm's performance to that of the optimal solution in hindsight (see e.g., \citeauthor{online_vrp_survey} for a survey). However, the worst-case theoretical nature of the measure makes it less suitable in more practical scenarios. 
Therefore, the online results are compared to the offline baselines while running them in a quasi-offline manner: at each planning step we re-run the offline algorithm including only opened jobs or customers. In CVRP this translates to solving an offline problem whenever the vehicle reaches the depot (all offline baselines assume the vehicle starts the route and the depot and therefore could not be used each time we reach a customer). In PMSP, we solve a new offline problem after each interval or when a machine becomes available.

To better understand the contribution of our approach and components of our solution, we evaluate SOLO in three ways: 
\begin{enumerate}
    \item Trained $\tilde Q$ alone (Offline part of SOLO)
    \item MCTS + Naive baseline as heuristic (Online part of SOLO)
    \item MCTS + Trained $\tilde Q$ as heuristic (Full SOLO solution)
\end{enumerate}
For the online problems, we run MCTS with rollout preemption, where in each rollout, the sampling of customers or jobs is stopped past a certain time. In addition, we use an action pruning mechanism to only evaluate a subset of the possible actions at each step.
Given the limited computing time, these techniques allow for a good balance between planning for the future on the one hand, but also seeing a broad view of the current state and possible actions. 
We train our policy, Q-net, which is based on $\tilde Q$, using a decaying learning rate with an initial value of $10^{-3}$. To allow for exploration the learning starts after $5{,}000$ steps and an $\varepsilon$ function is used, where the initial value of $\varepsilon$ is $1.0$ and it decays linearly over time. The training stops after $10^6$ steps and the model that achieves the best overall evaluation is saved and used. see the supplementary material for the full list of hyper-parameters used.

\vspace{0.5mm}
\textbf{CVRP:}
%\paragraph{CVRP}
In both online and offline cases, we consider the same setting as previous work (\citeauthor{chen2019learning, Kool2019ICLR, NIPS2018_nazari}) that consider three scenarios with $N \in \{20,50,100\}$ customers and vehicle capacities $C \in \{30,40,50\}$, respectively. Locations are sampled from a uniform distribution over the unit interval, and demands from a discrete uniform distribution over $\{ 0,\ldots, 10\}$.

In the online problem we introduce an arrival time in addition to the position and demand of each customer. Unlike the offline case, here the customer parameters are sampled from a Truncated Gaussian Mixture Model (TGMM). This sampling shows the strength of RL algorithms like ours, which take into account future customers by learning the distribution during training.

We compare our offline results to the following baselines: a trained Neural Rewriter model (\citeauthor{chen2019learning}), the two strongest problem heuristic baselines reported by \citeauthor{NIPS2018_nazari} (Savings and Sweep), Google’s OR-Tools, and the naive Distance Proportional baseline. 
In the latter baseline, each valid customer (whose demand is lower than the current capacity) is selected with probability proportional to its distance from the vehicle. 

\vspace{0.5mm}
\textbf{PMSP:}
%\paragraph{PMSP} 
For the offline problem we consider several cases with $80$ jobs, $3$ machines, and $5$ job classes.
We train a model on problems where job parameters are sampled from discrete uniform distributions (more details can be found in the supplementary material).
In the online setting we run two problems, both with $80$ jobs in expectation and $5$ job classes.
The first problem consists of $3$ machines and jobs arrive in $16$ intervals of length $130$ time units each. The second problem consists of $10$ machines and $60$ intervals with $10$ time units. We sample job processing times, setup times and job weights from discrete uniform distributions. Unlike the offline setting, here jobs arrive according to Poisson distributions, where the frequency of the $i$-th class is proportional to $1/i$. The problem parameters were chosen so as to not overload the machines in steady state, while maintaining relatively low levels of idleness.

We evaluate the offline problem on a public set of benchmark problems which uses the same ranges for job features \cite{Liao2012}.
We compare to the following baselines: a heuristic that is a variant of the weighted shortest processing times first heuristic (WSPT), the Neural Rewriter approach by \citeauthor{chen2019learning}, and the solution found after a fixed time using the IBM ILOG CPLEX CPOptimizer (V12.9.0) constraint programming suite.
More details about the baselines can be found in the supplementary material. In addition to the public benchmark we evaluate our solution on a smaller problem with 20 jobs, where the optimal solution is known.

\subsection{Experimental Results}
For each problem we train a Q-net model and run a comparison on 100 seeds between our method and the baselines mentioned above. 
%\paragraph{CVRP}
\begin{figure}[ht]
    \centering
    \includegraphics[width=0.75\columnwidth]{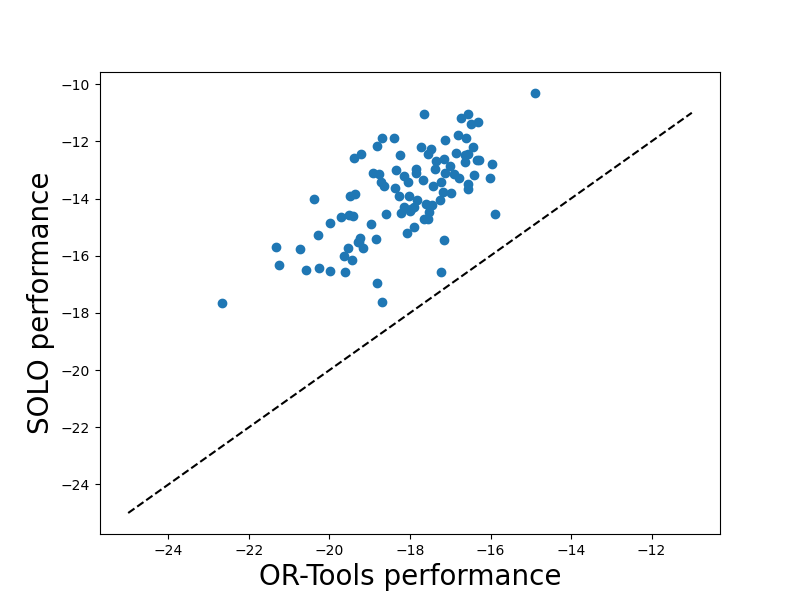}
    \caption{A performance comparison of SOLO and OR-Tools in the online CVRP50 settings.}
    \label{fig:scatter-dynamic-vrp-SOLO-vs-ortools}
\end{figure}

\vspace{0.5mm}
\textbf{CVRP:}
Both online and offline results can be found in Table~\ref{tab:CVRP-results}, more experiments are detailed in the appendix \cite{SOLOarxiv}. In the online problem SOLO outperforms all the other algorithms. In addition, despite SOLO being an algorithm for online problems, it reaches decent performance for the offline problem.  
When comparing SOLO to NeuralRewriter, the other learning algorithm, SOLO performs better and shows improved results.

We notice that the Savings algorithm achieves slightly better results than SOLO, which is expected since it is a problem specific heuristic and tailored to this problem. However, unlike SOLO, Savings does not extend to the online problem well and performs weakly when customers arrive over time.

When examining all the variants of our method we notice a number of interesting results. First, Q-net alone achieves good results compared to other algorithms, showing that the RL algorithm with the GNN architecture provides a good solution policy. Second, MCTS alone shows good results even when combined with a random heuristic, showing the value in online search algorithms and rollouts. Last, combining MCTS with our Q-net model achieves the best results showing the full strength of our method and necessity for both the offline learning and online search. 
Note that in all CVRP cases, SOLO with pruning did not show any improvement on SOLO alone and we decided not to include this in the plots.

In Figure~\ref{fig:scatter-dynamic-vrp-SOLO-vs-ortools} we see a comparison between SOLO and OR-Tools for the online problem with 50 customers. Each point above the dashed line is an instance where SOLO out-preformed OR-Tools. We see in the figure that all instances run in this evaluation are above the line and have better results when using SOLO.

\begin{figure}[ht]
    \centering
    \includegraphics[width=0.75\columnwidth]{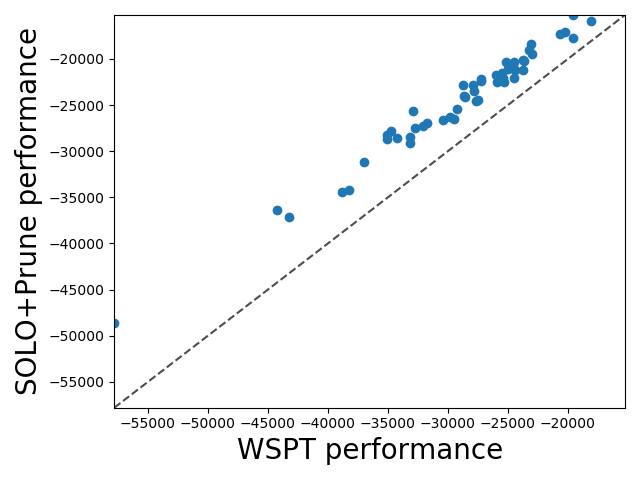}
    \caption{A  performance  comparison  of SOLO+Prune and WSPT in the online setting with $10$ machines and $50$ random instances.}
    \label{fig:scatter-dynamic-wspt-vs-mcts-q}
\end{figure}

\vspace{0.5mm}
\textbf{PMSP:}
%\paragraph{PMSP} 
We evaluate the model trained on $20$ jobs on a public set of offline benchmarks where the optimal solution quality is known \cite{KRAMER2019}. 
The results in Table \ref{tab:pms_results} show our approach performs mostly on par with CPLEX. In addition, our method out-performs the other learning algorithm (Neural Rewriter) and shows good results when evaluating on $20$ and $80$ jobs.
Similar to CVRP, MCTS improves the results achieved by Q-net alone, and shows that adding an online search algorithm to the offline learned model improves results. 
The scatter plot in Figure \ref{fig:scatter-dynamic-wspt-vs-mcts-q} shows that in the online scheduling problem, our approach yields better solutions than WSPT in all problem instances.

\begin{figure}[ht]
    \centering
    \includegraphics[width=0.75\columnwidth]{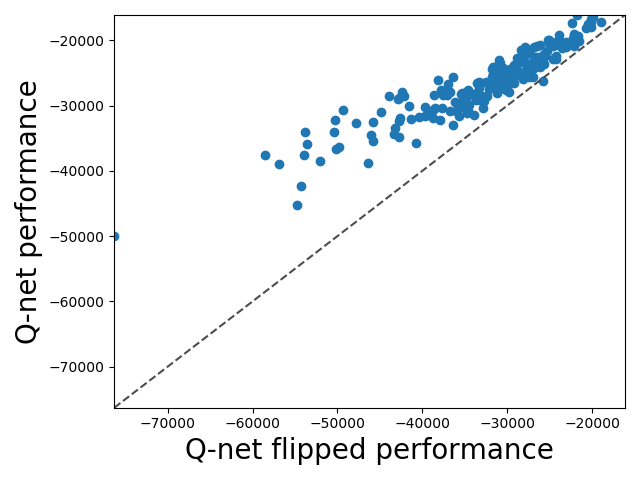}
    \caption{Performance comparison of Q-net to its performance in the online setting with ``flipped classes''. $10$ machines, $60$ arrival intervals, interval durations of $10$, and $200$ random instances.}
    \label{fig:dynamic-m10-swap}
\end{figure}
To analyze whether the Q-net model actually learns to take advantage of the arrival distribution, we apply it in a scheduling setting where we flip the class identity of observed jobs from $i$ to $5-i$. The results shown in Figure \ref{fig:dynamic-m10-swap} indicate that the model is indeed optimized for the arrival distribution used during training.

\begin{table}[ht]
 \centering\fontsize{8.8}{11}\selectfont
\centering 
% \textbf{Offline CVRP}\\
\begin{tabular}{lll}
\multicolumn{3}{c}{\textbf{Offline CVRP}} \\
  &  \multicolumn{1}{c}{\textbf{20}} & \multicolumn{1}{c}{\textbf{100}} \\
Uniform-Random[UR] & -13.21(107.51\%)  & -58.84 (230.13\%) \\
Distance[D] & -10.43 (63.65\%)& -47.59 (167.38\%)   \\
Savings & -6.35 (-1.04\%) & \textbf{-16.51 (-7.94\%)}   \\
Sweep & -8.89 (39.33\%)  & -28.24 (58.11\%)   \\
OR-Tools & -6.42 (0.00\%)  & -17.96 (0.00\%)   \\
NeuralRewriter & -6.95 (8.48\%)  & -19.45 (8.57\%) \\
\cline{1-3} \\
\textbf{Q-Net} & -6.84 (6.59\%)  & -19.27 (7.62\%)  \\
\textbf{MCTS+UR} & -7.65 (19.45\%)  & -46.34 (160.11\%)   \\
\textbf{MCTS+D} & -7.15 (12.01\%)  & -44.00 (147.44\%)  \\
\textbf{SOLO} & \textbf{-6.21 (-3.18\%)}  & -17.68 (-1.24\%)  \\ 
\\
\multicolumn{3}{c}{\textbf{Online CVRP}} \\
& \multicolumn{1}{c}{\textbf{20}} & \multicolumn{1}{c}{\textbf{100}} \\ 
Uniform-Random[UR] & -12.72 (31.67\%)  & -52.73 (108.06\%)    \\
Distance[D] & -9.75 (0.76\%)  & -33.65 (32.72\%)    \\
Savings & -9.90 (0.51\%)  & -25.15 (-0.90\%)    \\
Sweep & -11.16 (13.73\%)  & -29.52 (16.16\%)   \\
OR-Tools & -9.86 (0.00\%)  & -25.40 (0.00\%)    \\
NeuralRewriter & -10.00 (1.56\%)  & -25.85 (1.90\%)   \\ 
\cline{1-3} \\
\textbf{Q-net} & -8.79 (-9.76\%) & -26.80 (5.70\%)   \\
\textbf{MCTS+UR} & -7.80 (-20.27\%) & -28.72 (12.98\%) \\
\textbf{MCTS+D} & -6.78 (-30.84\%)  & -25.58 (0.78\%)  \\
\textbf{SOLO} & \textbf{-6.63 (-32.38\%)} & \textbf{-24.80 (-2.28\%)} 
\end{tabular}%
% }
\caption{Offline and Online CVRP results. Each cell contains the average cost and the fractional improvements over OR-Tools (negative numbers are better than OR-Tools). Best results and our methods are marked in bold}
\label{tab:CVRP-results}
\end{table}

\begin{table}[!hbt]
\centering
% \fontsize{9}{11}\selectfont
\resizebox{\columnwidth}{!}{%
\begin{tabular}{lll}
\multicolumn{3}{c}{\textbf{Offline PMSP}} \\
 & \multicolumn{1}{c}{\textbf{liao 20}} & \multicolumn{1}{c}{\textbf{liao 80}} \\
WSPT & -16570.16 (5.82\%) & -182357.02 (4.15\%) \\
CPLEX & -15658.46 (0\%) & -175084.88 (0\%) \\
NeuralRewriter & -16540.28 (5.63\%) & -182450.02 (4.21\%) \\ \hline
\\
\textbf{Q-net} & -15906.32 (1.58\%) & -178444.74 (1.92\%) \\
\textbf{MCTS+WSPT} & -15876.88 (1.39\%) & -176439.74 (0.77\%) \\
\textbf{SOLO} & -15695.94 (0.24\%) & -175524.34 (0.25\%) \\
\textbf{SOLO+Prune} & \textbf{-15683.46 (0.16\%)} & \textbf{-175164.58 (0.05\%)} \\
optimal & -15628.68 (-0.19\%) &  \\
\\
\multicolumn{3}{c}{\textbf{Online PMSP}} \\
 & \multicolumn{1}{c}{\textbf{3 machines}} & \multicolumn{1}{c}{\textbf{10 machines}} \\ 
WSPT & -40601.34 (15.04\%) & -29102.5 (18.87\%) \\
CPLEX & -35294.38 (0\%) & -24481.9 (0\%) \\
NeuralRewriter & -38575.78 (9.3\%) & -27350.68 (11.72\%) \\ \hline
\\
\textbf{Q-net} & -37386.9 (5.93\%) & -26031.5 (6.33\%) \\
\textbf{MCTS+WSPT} & -35489.56 (0.55\%) & -24724.76 (0.99\%) \\
\textbf{SOLO} & -35434.46 (0.4\%) & -24747.38 (1.08\%) \\
\textbf{SOLO+Prune} & \textbf{-35280.2 (-0.04\%)} & \textbf{-24655.42 (0.71\%)}
\end{tabular}%
}
\caption{Scheduling results for all problem variants. Each cell includes the average cost on 50 seeds and the fractional improvement of each method compared to CPLEX (negative numbers are better than CPLEX).}
\label{tab:pms_results}
\end{table}

\section{Conclusion and Discussion}
\label{sec:conclusions}
We presented a deep RL approach to combinatorial optimization. Our method incorporates graph-based representation of the state-action space into Deep Q-learning to learn an input-size agnostic policy, and we further combine it with MCTS to significantly improve performance of our approach. 
% The Q-net automatically takes the input distribution into account during training as it interacts with the simulator. The combination with MCTS can even guarantee anytime optimality, e.g., when using UCT.
We evaluated these contributions on two combinatorial optimization problems, PMSP and CVRP.
Our pure Q-net agent, which provides near-instantanous action selection, outperforms popular heuristics. Conversely, our combined approach, SOLO, makes full use of any available time for deliberation thanks to its anytime nature, effectively decreasing the gap to dedicated combinatorial optimization solvers.
As a future work direction we propose to investigate explicitly learning the arrival distribution
of the online combinatorial optimization problem during training. Another direction is to explore alternative graph representation for the states.

% \textbf{The advantage of learning the input distribution}
% % \paragraph{The advantage of learning the input distribution}
% Many of the known baseline approaches for online variants are based on simple modifications that disregard the underlying arrival distributions. One would expect that a method adept at learning the arrival distribution, either explicitly or implicitly (as is the case for Q values) would improve upon such naive, myopic approaches~\cite{garg08}. In fact, one can prove that for both CVRP and scheduling, the competitive ratio between an online and an offline algorithm in an online setting can be arbitrarily high. \at{things from the appx can be added here as a discussion and in relation to the future work...}

%\input{acknowledgments}

\clearpage

% \fontsize{9.8pt}{10.8pt} \selectfont
\bibliography{bib2}

% \clearpage
\appendix
\section{Appendix}
\paragraph{Organization} The appendix is organized as follows: Section \ref{app:baselines} presents the different baselines used for comparisons. Section \ref{app:training} shows a detailed explanation of the Q-net architecture and training. Section \ref{app:evaluations} includes additional experiments and deeper explanations of results shown in Section \ref{sec:empirical_evalutaion}. Finally, Section \ref{app:inputdistribution} has a theoretical example showing the added value of algorithms that learn the online distribution and use this knowledge to better solve the online problem.

In addition to this document, we supply training and evaluation source code. See the \texttt{readme.md} file in the supplementary material zip file.

%\section{Benchmark Details}\label{app:benchmark}
\chr{not sure we need this whole section since we have a lot of this in the main part of the paper}
\jo{I agree. This section is already described in good detail in the main body of the paper. I'm commenting it out.}

\section{Baseline Details}\label{app:baselines}
\paragraph{CVRP}
All our baselines share the following technique: In order to be able to apply them to online problems, we let them recompute their route every time the agent visits the depot. Conceptually, it would be possible to recompute routes at every time step, which would allow the agents to react faster to new arrivals. However, this would require substantial changes to the baseline implementations, since they all assume that the agent is at the depot when a route is computed.

We use the savings baseline implementation from the VeRyPy project \cite{verypy}. Note that although this implementation differs from the original one by \citealp{NIPS2018_nazari} (and also used by, e.g., the Neural Rewriter paper \citealp{chen2019learning}), both claim to implement the same Clarke-Wright Savings algorithm. \cite{ClarkeWright1964}. The superior performance we report must therefore stem from differences in the implementation. For example, a performance of $12.25$ is reported for the best savings variant on CVRP problems of size $50$ in related work. The VeRyPy implementation achieves a value of $11.03$.

For the Neural Rewriter baseline, we used the repository version. We trained models for all settings for 24 hours with the default hyperparameters. Note that the difference in performance we report on the CVRP problem compared to the original paper stems from the fact that we train Neural Rewriter models for 24h in order to try to give Neural Rewriter and our approach the same amount of training time. Training our models is always completed after at most 20 hours. For the online problems, we trained on offline problems of corresponding size where the customer positions were drawn from the Gaussian mixture distributions defined for the online settings.

The OR tools baseline is the one implemented by \citealp{NIPS2018_nazari}. One difference is that we allow the policy to reoptimize its route every time the agent visits the depot. This is often an advantage, since it is easier to optimize a smaller remaining subproblem. The solver was executed with a 10 second time limit per optimization.

\paragraph{PMSP}
The WSPT variant we use selects, at a given state, job and machine $(j, m)$ such that $(f(s_m, s_j) + p_j) / w_j$ is minimal, where $s_m$ is the current class of machine $m$, and $p_j$ and $w_j$ are the processing time and weight of $j$, respectively. 

In order to apply the Neural Rewriter approach to parallel machine scheduling we modified the authors' implementation for the scheduling problem. We represent each schedule as a single list, starting with an entry corresponding to the first machine, then entries for the jobs scheduled to the first machine, an entry for the second machine, entries for the jobs scheduled to the second machine, and so on. Initial schedules are created by applying the WSPT rule. We also tried to create initial schedules by uniformly choosing actions, with similar results. Schedules can be manipulated by swapping any two jobs. We use the same features as for our approach, with the addition of, for each job, the time it is started, the time the setup change is completed, and the time the job itself is completed. We tuned the hyperparameters with a grid search around the provided default hyperparameters. We allowed Neural Rewriter to take $50$ rewriting steps and return the best schedule. We trained models for 24 hours. The training instances are sampled during training according to the ranges defined above. Since it is not possible to train the Neural Rewriter baseline directly on the online settings, we use the model trained on the $20$ job offline setting instead in the online settings.

CPOptimizer was executed with a 10 second time limit on an Intel(R) Core(TM) i9-9820X CPU @ 3.30GHz with 64GB of RAM.

\section{DQN and Q-net Details}\label{app:training}
\paragraph{Network Architecture Details}

\begin{table}[t]
    \resizebox{1.0\columnwidth}{!}{%
    \begin{tabular}{*6c}
        %\toprule
        \textbf{Problem} & \textbf{Network} & \textbf{Embedding} & \textbf{GNN} & \textbf{Output} & \textbf{N-passes}  \\ 
        %\midrule
        PMS   & $\phi_e$, $\phi_n$ & $32$  & $32 \quad 32 \quad 32$ & $1$  &  \multirow{2}{*}{4} \\ 
        & $\phi_g$ & $32$  & $32 \quad 32 \quad 32$ & $1$  & \\
         CVRP   & $\phi_e$, $\phi_n$ & $256$  & $256 \quad 256 \quad 256$ & $1$  &  2\\
        & $\phi_g$ & $256$  & $256 \quad 256 \quad 256$ & $2$  & \\
    \end{tabular}%
    }
    \caption{Output dimensions of fully connected layers and number of message passes for edge $\phi_e$, node $\phi_n$ and global $\phi_g$ networks}
    \label{tab:totalParams}
\end{table}

As mentioned in Section \ref{sec:implementation_details} we use an embedding - encoder - decoder model where the embedding is a simple feed forward model. The encoder and decoder models are both GNN's with the same message-passing mechanism introduced by \citealp{battaglia2018relational}. 
Table~\ref{tab:totalParams} summarizes the dimensions used in each fully connected layer and the number of passes through the GNN message passing section of the model.

Our GNN architecture does not effect the graph structure but only the nodes, edges and globali features.
The final edge features are the Q-values of each action (in CVRP $e_{i, j}$ represents choosing customer $j$ and in PSMP it represents scheduling job $j$ on machine $i$).
The global feature $w$ represents the Q-value of the \texttt{noop} action\footnote{Note that $w$ has dimensionality 2 for CVRP because of dueling, see Section \ref{app:dqn-details}}.

\paragraph{DQN details}
\label{app:dqn-details}
Our DQN implementation includes various improvements that were suggested by \citealt{hessel2018rainbow} such as dueling, double Q, delayed target network updates, prioritized replay and multi-step reward estimation.
Once implementing all of the improvements we ran a hyper-parameter tuning algorithm (based on successive halving \citealp{kumar2018parallel}) and found that some of the improvements did not demonstrate any increase in performance. 

\begin{table}[ht]
    \resizebox{1.0\columnwidth}{!}{%
    \begin{tabular}{l*4l}
        %\toprule
        \multirow{2}{*}{\textbf{Parameter}} & \multicolumn{2}{c}{\textbf{PMS}} & \multicolumn{2}{c}{\textbf{CVRP}} \\ 
        & Offline & Online & Offline & Online \\
        %\midrule
        total time steps                     & $1e^6$ & $1e^6$ & $1e^6$ & $1e^6$ \\
        target network update frequency     & $5e^3$ & $5e^3$ & $5e^3$ & $5e^3$ \\
        initial random steps                & $5e^3$ & $5e^3$ & $5e^3$ & $5e^3$ \\
        learning starts $t$                 & $5e^3$ & $5e^3$ & $5e^3$ & $5e^3$ \\
        train frequency                     & $1$ & $1$ & $1$ & $1$ \\
        discount $\gamma$                   & $1.0$ & $0.9$ & $1.0$ & $1.0$ \\
        batch size                          & $32$ & $32$ & $128$ & $128$ \\
        replay buffer size                  & $5e^3$ & $5e^3$ & $5e^3$ & $5e^3$ \\
        prioritized replay $\alpha$         & $0.0$ & $0.0$ & $25e^{-3}$ & $25e^{-3}$ \\
        prioritized replay $\beta_0$        & $0.4$ & $0.4$ & $0.4$ & $0.4$ \\
        prioritized replay $e$              & $1e^{-6}$ & $1e^{-6}$ & $1e^{-6}$ & $1e^{-6}$ \\
        exploration fraction                & $0.3$ & $0.3$ & $0.1$ & $0.1$ \\
        exploration final                   & $0.1$ & $1e^{-4}$ & $1e^{-4}$ & $1e^{-4}$ \\
        learning rate                       & $1e^{-3}$ & $1e^{-3}$ & $1e^{-3}$ & $1e^{-3}$ \\
        learning rate $e$                   & $0.0$ & $0.0$ & $0.1$ & $0.1$ \\
        gradient norm clipping              & $2e^2$ & $2e^2$ & $2e^2$ & $2e^2$ \\
        double $Q$                          & \texttt{false} & \texttt{false} & \texttt{false} & \texttt{false} \\
        dueling                             & \texttt{false} & \texttt{false} & \texttt{true} & \texttt{true} \\
        $n$ step reward estimation          & $1$ & $1$ & $1$ & $1$ \\
        %\bottomrule
    \end{tabular}%
    }
    \caption{Training DQN hyper parameters used for all problems}\label{tab:DQNParams}
\end{table}
In addition, we found that, although not warranted for theoretical reasons, introducing a discount factor helped numeric convergence in the online PMSP problem. We find that these scenarios demonstrate a high variance in the returns and make it harder for the training process to efficiently learn the Q values.
The final hyper-parameters for the two benchmark problems are summarized in Table \ref{tab:DQNParams}.
Note that a different Q-net policy was trained for each problem size but all problems used the same DQN hyper-parameters.

\begin{table}[t]
    \resizebox{1.0\columnwidth}{!}{%
    \begin{tabular}{lll}%
        \textbf{Problem} & \textbf{GPUh} & \textbf{Resources} \\
        CVRP Offline size $20$  & $13.08$ & \multirow{2}{*}{\shortstack[l]{$18$-core Intel \\ Xeon Gold $6150$ @ $2.70$ GHz \\ Nvidia Tesla V$100$ ($16$ GB)}} \\ 
        CVRP Offline size $50$  & $14.68$ \\
        CVRP Offline size $100$ & $20.46$ \\[5pt]
        \hline \\
        CVRP Online size $20$   & $5.08$ & \multirow{2}{*}{\shortstack[l]{20-core Intel i9-10900X @ 3.70GHz \\ Nvidia GeForce RTX 2080 (11 GB)}} \\ 
        CVRP Online size $50$   & $17.05$ \\
        CVRP Online size $100$  & $7.38$ \\[5pt]
        \hline \\
        PMS Offline size $20$   & $14.3$ & \multirow{2}{*}{\shortstack[l]{$18$-core Intel\\ Xeon Gold $6150$ @ $2.70$ GHz \\ Nvidia Tesla V$100$ ($16$ GB)}}\\
        PMS Offline size $80$   & $21.83$ \\
        PMS Online $3$ machines & $6.9$ \\
        PMS Online $10$ machines& $10.65$ \\
    \end{tabular}%
    }
    \caption{Training time until the best model for the benchmark problems reported in GPU hours.}
    \label{tab:training_time}
\end{table}
% ($3.70$ GHz max. Turbo)

Each training experiment was limited to 24 GPU hours. 
Table~\ref{tab:training_time} summarizes the computational resources and training time for each problem. The time noted is the number of hours needed to achieve the best model performance (the training once models stopped improving).

The CVRP problem was able to train faster and therefore we were able to increase the batch size and enable prioritized replay buffer (this increases computational complexity of sampling and therefore can only be used when training is fast enough). In addition, we were able to increase the amount of hidden units in each fully connected layer of the GNN architecture. (specific numbers can be found in Table~\ref{tab:DQNParams})

\section{Evaluation Details}\label{app:evaluations}

\begin{table}[t]
    \small
    \begin{tabular}{l*5c}
        \multicolumn{1}{l}{Distribution} & \textbf{$\mu$} & \textbf{$\sigma$} & \textbf{$w$} & \textbf{$a$} & \textbf{$b$} \\ \\
        \multirow{2}{*}{position $(x, y)$}  & $(0.25, 0.25)$           & $(0.1, 0.1)$  & $0.5$  & $0$ & $1$  \\
         & $(0.75, 0.75)$   & $(0.1, 0.1)$  & $0.5$  & $0$ & $1$  \\
        \hline \\
        \multirow{2}{*}{time $t$}   & \multicolumn{1}{c}{5}  & \multicolumn{1}{c}{3} & 0.33 & 0 & 40 \\
                                    & \multicolumn{1}{c}{20} & \multicolumn{1}{c}{3} & 0.33 & 0 & 40 \\
                                    & \multicolumn{1}{c}{40} & \multicolumn{1}{c}{3} & 0.33 & 0 & 40 \\
    \end{tabular}{}
    \caption{Truncated Gaussian components of Mixture Model distribution for the Online CVRP. $\mu, \sigma, \omega, a$ and $b$ stand for mean, standard deviation, weight, minimum value and maximum value of each truncated Gaussian distribution correspondingly.} \label{tab:online_distribution}
\end{table}

\paragraph{CVRP Online distribution}
The distribution chosen for the customer times in the online problem is a Truncated Gaussian Mixture distribution. Each distribution is represented by a finite number of sets: $(\mu, \sigma, \omega, a, b)$ noting the mean, standard deviation, weight, minimum value and maximum value of each truncated Gaussian distribution. 
In addition the position of customers is sampled from a $2d$ Truncated Gaussian Mixture distribution where $(x, y)$ are sampled together.
For details on the specific distribution values see Table \ref{tab:online_distribution}.

\begin{figure}[ht]
    \centering
    \includegraphics[width=0.75\columnwidth]{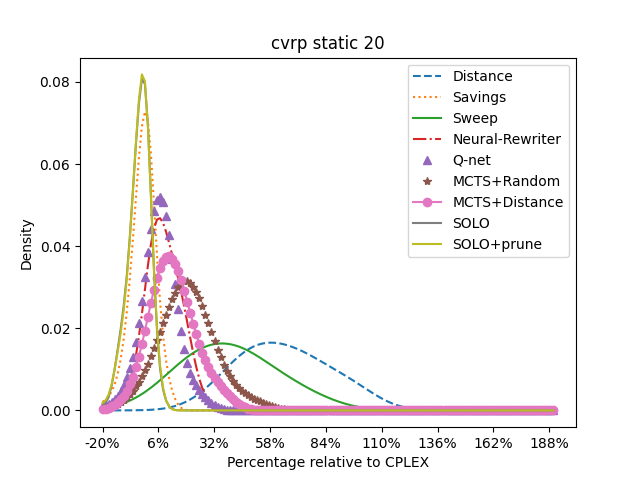}
    \caption{Offline CVRP with 20 customers.}
    \label{fig:CVRP_offline_20}
\end{figure}

\begin{figure}[ht]
    \centering
    \includegraphics[width=0.75\columnwidth]{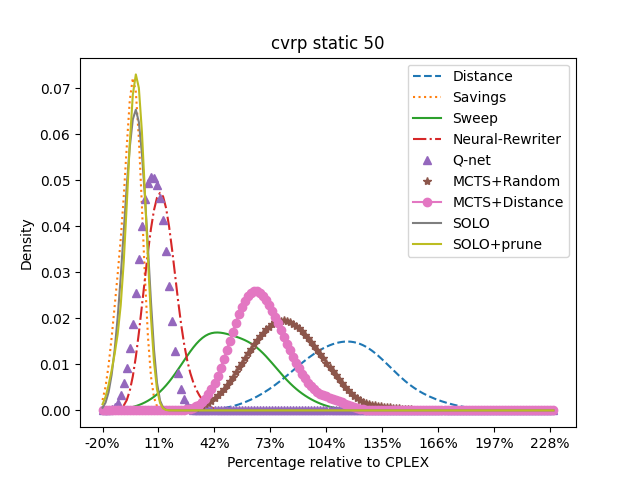}
    \caption{Offline CVRP with 50 customers.}
    \label{fig:CVRP_offline_50}
\end{figure}

\begin{figure}[ht]
    \centering
    \includegraphics[width=0.75\columnwidth]{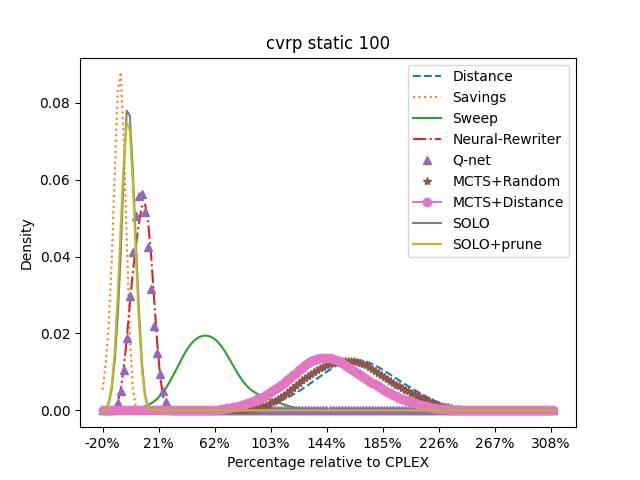}
    \caption{Offline CVRP with 100 customers.}
    \label{fig:CVRP_offline_100}
\end{figure}
\begin{figure}[ht]
    \centering
    \includegraphics[width=0.75\columnwidth]{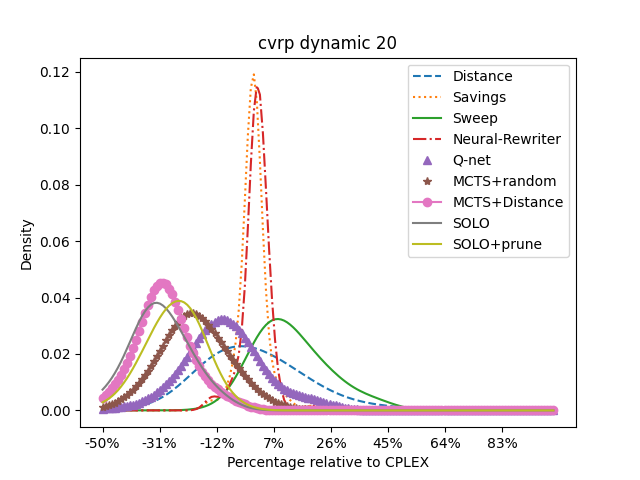}
    \caption{Online CVRP with 20 customers.}
    \label{fig:CVRP_online_20}
\end{figure}

\begin{figure}[ht]
    \centering
    \includegraphics[width=0.75\columnwidth]{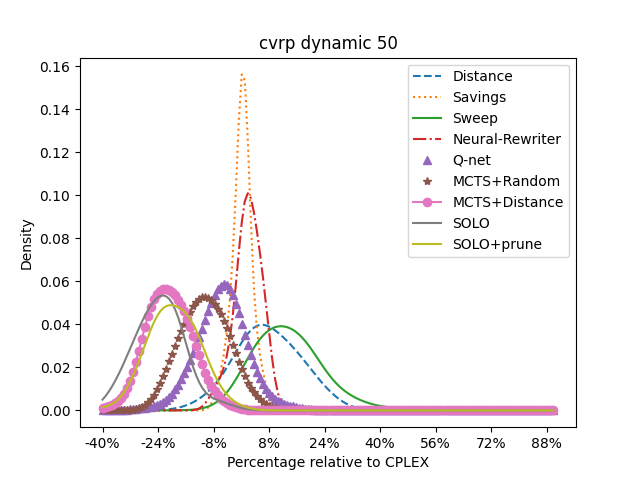}
    \caption{Online CVRP with 50 customers.}
    \label{fig:CVRP_online_50}
\end{figure}

\begin{figure}[ht]
    \centering
    \includegraphics[width=0.75\columnwidth]{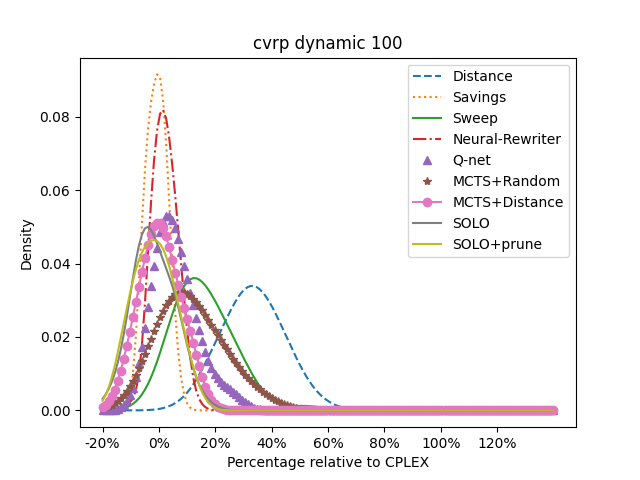}
    \caption{Online CVRP with 100 customers.}
    \label{fig:CVRP_online_100}
\end{figure}

In Figures~\ref{fig:CVRP_offline_20}-\ref{fig:CVRP_online_100} a deeper comparison between SOLO and other baselines can be found for the CVRP online and offline problems. Each line in the figures represents the distribution of relative rewards achieved by the baselines compared to the OR-Tools baseline (baselines with a negative mean value reached better mean results compared to OR-Tools). 
% These figures show the full distribution of the results summarized in Section \ref{sec:empirical_evalutaion}, Table \ref{tab:CVRP-results}. 
These figures are summarized in table \ref{tab:CVRP-results_appx}.
We see that in most cases our full method out preforms other algorithms and achieves good results.

\paragraph{PMSP}
\begin{figure}[!ht]
    \centering
    \includegraphics[width=0.75\columnwidth]{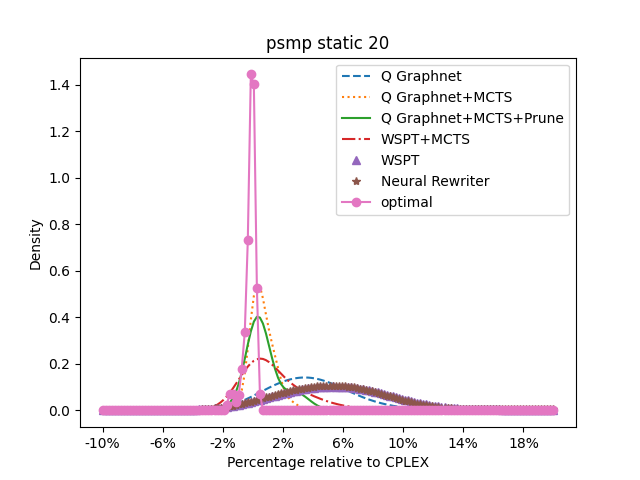}
    \caption{Offline PSMP with 20 jobs. Distribution of evaluations compared to CPLEX.}
    \label{fig:PSMP_offline_20}
\end{figure}
\begin{figure}[!ht]
    \centering
    \includegraphics[width=0.75\columnwidth]{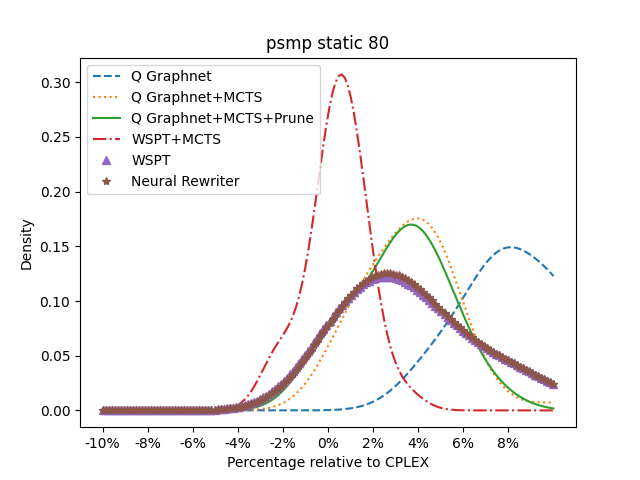}
    \caption{Offline PSMP with 80 jobs. Distribution of evaluations compared to CPLEX.}
    \label{fig:PSMP_offline_80}
\end{figure}
\begin{figure}[!ht]
    \centering
    \includegraphics[width=0.75\columnwidth]{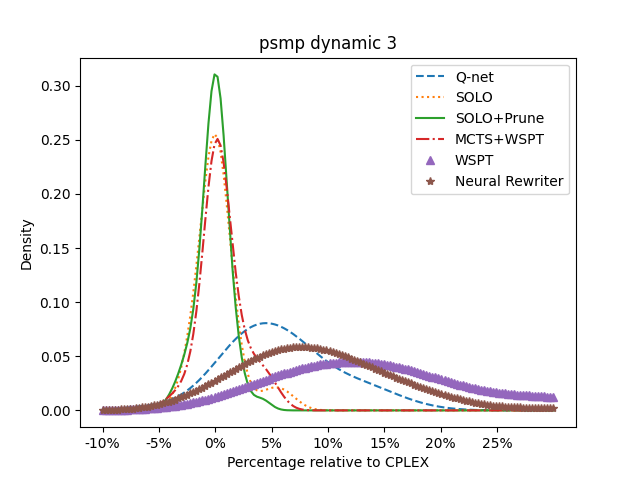}
    \caption{Online PSMP with 3 machines.}
    \label{fig:PSMP_online_3}
\end{figure}
\begin{figure}[!ht]
    \centering
    \includegraphics[width=0.75\columnwidth]{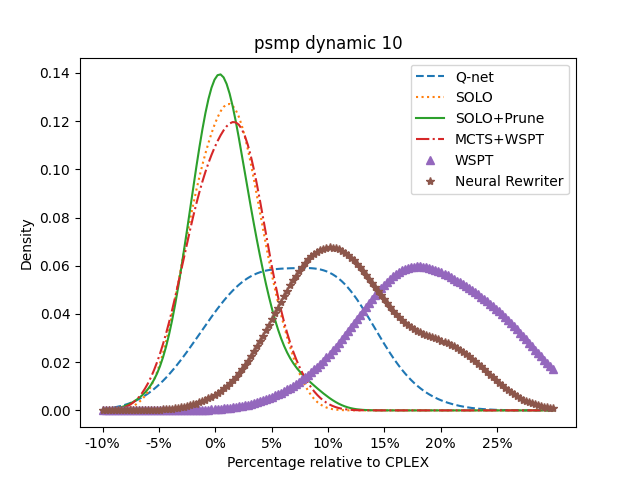}
    \caption{Online PSMP with 10 machines.}
    \label{fig:PSMP_online_10}
\end{figure}
For all problems, we sample the job processing times, setup times, and job weights from the discrete uniform distributions: $U[1, 100]$, $U[1, 50]$ and $U[1, 10]$, respectively.
Similar to the figures shown above, Figures~\ref{fig:PSMP_offline_20} - \ref{fig:PSMP_online_10} depict our comparisons between the SOLO and the various baselines, relative to the CPLEX baseline, for the PSMP Offline and Online problems (negative average indicates better results).

\subsection{Runtime}
As mentioned in Section \ref{sec:empirical_evalutaion} all benchmarks including SOLO were run using a timeout of $10$ seconds. In SOLO, the time given limits the amount of rollouts MCTS can preform for each state. We can compute the worst-case runtime in seconds as $n_d \times 10$ with $n_d$ being the number of decision points.
For the Offline CVRP with disabled \texttt{noop} action (i.e., the agent is forced to move to some customer after it arrived to the depot) worst-case $n_d$ is computed as $2 \times s_p$, where $s_p$ is the problem size. This corresponds to the sub-optimal but yet acceptable strategy of routing back to the depot after every customer.
For the Online setting worst-case $n_d$ equals to $3 \times s_p$ and corresponds to waiting at the depot until all customers become visible and then following the sub-optimal offline strategy described above.
Therefore the worst-case runtime will be $200$, $500$ and $1000$ seconds for the problem sizes $20$, $50$ and $100$ respectivly .
Worst-case runtime of the Offline PMS will add up to $200$ and $800$ seconds for problems of size $20$ and $80$.
For the Offline PMS $n_d$ is equal to $s_p$ since the \texttt{noop} action is disabled, meaning that at every decision point SOLO has to schedule one of the available jobs.
For both the online PMS problems with different numbers of machines, $n_d$ is equal to the expected number of arriving jobs that is set to $80$.
% That will result in $800$ seconds of worst-case runtime.

\section{The Advantage of Learning the Input Distribution}
\label{app:inputdistribution}
%A common ingredient in the online variants of combinatorial problems is a distribution over possible input streams. For example, in dynamic scheduling, the input stream consists of sequences of jobs, and in the capacitated vehicle routing problem the customers arrive one at a time in a non-deterministic fashion.

A well-studied problem in the field of online algorithms is the limiting effect of not having information about the arrival distribution (e.g., \citealp{garg08}). In contrast, modern reinforcement learning algorithms often demonstrate a capacity to incorporate, either explicitly or implicitly, knowledge of the distribution (e.g., in their estimates of the Q values). In this context, we pose the following question: 

% \medskip

\textit{Can an efficient optimal, or approximately optimal offline algorithm be naively adapted to the online version of the problem without regard for the arrivals distribution?}

It is not hard to show that such adaptations generally do not preserve their offline optimality properties. To see that, consider the following examples for the two problems studied in this paper:

\vspace{0.5mm}
\textbf{Online Capacitated Vehicle Routing:}
% \paragraph{Online Capacitated Vehicle Routing:} 
In this case, consider $n$ arrivals from two distant locations $p_1$ and $p_2$ between time $0$ and the horizon $h$. Each customer arrives at time $t \sim U[0, h]$ and is positioned at either location with probability $\frac{1}{2}$, each. It can be shown that, in expectation, an offline optimal algorithm that is run in an online fashion would make $\Omega(n)$ travels between the two locations, whereas the algorithm that first handles customers at location $p_1$ and then moves on to location $p_2$ will make $O(1)$ such trips. As before, the expected ratio is thus unbounded.
%\fr{@Joel, I don't think we can get around specifying the vehicle's speed in this example. But I don't think we can just assume it is fast enough to go between customers before the next customer arrives, because there is no lower bound on the time between customer arrivals.}\jo{@Felix , in \textit{expectation} (but not in all instantiation), the arrivals come at even intervals of time length $\frac{h}{n}$. Suppose the time it takes the vehicle to commute from one location to the other is half than that. Then in expectation, the vehicle will have $O(1)$ customer waiting (or more likely, single customer) at each decision point. No?} \fr{Yes, I don't question the validity of the argument. I guess I'm worried we ask too much of the reader here, since we require the reader to either believe or try to do the math themselves. Especially after we start the sentence with "Clearly".} \jo{Fair enough. I changed the wording in the beginning to ``It can be shown that,'', which is more prudent.}

\vspace{0.5mm}
\textbf{Online Scheduling:}
Consider a simple distribution in which there are two job classes $s_1$ and $s_2$, where each incoming job belongs to either class with probability $\frac{1}{2}$. All weights and processing times are $\epsilon$. The setup time incurred by switching from class $s_1$ to $s_2$ is $1$, and the setup time incurred by switching from class $s_2$ to $s_1$ is again, $\epsilon$. Now consider a minimal instantiation of the problem, in which there are two machines, and two batches of size $2$, each. Then with probability $\frac{3}{16}$, the first batch will not contain class $s_2$ jobs, whereas the second batch will. In this case, an optimal offline distribution-agnostic algorithm will assign the two jobs to separate machines. On the other hand, a
reasonable algorithm with access to the distribution will reserve a machine for future class $s_2$ jobs, due to its prohibitive setup costs, and put all $s_1$ jobs on the other machine. Since such instances occur with 
constant probability, it is easy to see that the online algorithm has a TWCT objective value of 
$O(\epsilon)$, while the offline algorithm's objective value will be $1 + O(\epsilon)$. Hence, the expected ratio of costs of the two algorithms will be unbounded as $\epsilon$ approaches $0$.

\begin{table*}[!ht]
\centering \textbf{Offline CVRP}\\
\begin{tabular}{llll}
& \multicolumn{1}{c}{\textbf{20}} & \multicolumn{1}{c}{\textbf{50}} & \multicolumn{1}{c}{\textbf{100}} \\
Uniform-Random[UR] & -13.21 (107.51\%) & -30.58 (167.52\%) & -58.84 (230.13\%)  \\
Distance[D] & -10.43 (63.65\%) & -24.36 (113.39\%) & -47.59 (167.38\%)   \\
Savings & -6.35 (-1.04\%) & \textbf{-11.03 (-3.98\%)} & \textbf{-16.51 (-7.94\%)} \\
Sweep & -8.89 (39.33\%) & -17.23 (49.94\%) & -28.24 (58.11\%) \\
OR-Tools & -6.42 (0.00\%) & -11.50 (0.00\%) & -17.96 (0.00\%)   \\
NeuralRewriter & -6.95 (8.48\%) & -12.83 (11.83\%) & -19.45 (8.57\%) \\
\cline{1-4} \\
\textbf{Q-Net} & -6.84 (6.59\%) & -12.33 (7.37\%) & -19.27 (7.62\%)  \\
\textbf{MCTS+UR} & -7.65 (19.45\%) & -20.72 (81.08\%) & -46.34 (160.11\%)   \\
\textbf{MCTS+D} & -7.15 (12.01\%) & -18.91 (65.04\%) & -44.00 (147.44\%)  \\
\textbf{SOLO} & \textbf{-6.21 (-3.18\%)} & -11.25 (-1.98\%) & -17.68 (-1.24\%)  \\ 
\\
\multicolumn{4}{c}{\textbf{Online CVRP}} \\
& \multicolumn{1}{c}{\textbf{20}} & \multicolumn{1}{c}{\textbf{50}} & \multicolumn{1}{c}{\textbf{100}} \\ 
Uniform-Random[UR] & -12.72 (31.67\%) & -28.00 (54.25\%) & -52.73 (108.06\%)    \\
Distance[D] & -9.75 (0.76\%) & -19.70 (8.48\%) & -33.65 (32.72\%)    \\
Savings & -9.90 (0.51\%) & -18.19 (0.07\%) & -25.15 (-0.90\%)    \\
Sweep & -11.16 (13.73\%) & -20.75 (14.27\%) & -29.52 (16.16\%)   \\
OR-Tools & -9.86 (0.00\%) & -18.18 (0.00\%) & -25.40 (0.00\%)    \\
NeuralRewriter & -10.00 (1.56\%) & -18.54 (2.00\%) & -25.85 (1.90\%)   \\ 
\cline{1-4} \\
\textbf{Q-net} & -8.79 (-9.76\%) & -17.26 (-4.99\%) & -26.80 (5.70\%)   \\
\textbf{MCTS+UR} & -7.80 (-20.27\%) & -16.54 (-9.06\%) & -28.72 (12.98\%) \\
\textbf{MCTS+D} & -6.78 (-30.84\%) & -14.59 (-19.74\%) & -25.58 (0.78\%)  \\
\textbf{SOLO} & \textbf{-6.63 (-32.38\%)} & \textbf{-14.03 (-22.82\%)} & \textbf{-24.80 (-2.28\%)}
\end{tabular}%
% }
\caption{Offline and Online CVRP results. Each cell contains the average cost and the fractional improvements over OR-Tools (negative numbers are better than OR-Tools). Best results and our methods are marked in bold}
\label{tab:CVRP-results_appx}
\end{table*}

\end{document}